\documentclass{isprs} 
\usepackage{subfigure}
\usepackage{setspace}
\usepackage{geometry} 
\usepackage{epstopdf}
\usepackage{microtype}
\usepackage[labelsep=period]{caption}  
\usepackage[hang]{footmisc}
\usepackage{multirow}
\usepackage{xcolor}
\usepackage{graphicx}
\usepackage{lmodern}
\usepackage{setspace}
\usepackage{amsmath}
\usepackage{adjustbox}
\usepackage{booktabs} 
\usepackage{cuted} 
\usepackage{placeins}
\usepackage{tabularx}
\usepackage{ulem}
\usepackage{amssymb}
\usepackage{babel}
\usepackage{url}

\definecolor{Gcolor}{RGB}{220, 38, 38}     
\definecolor{Rcolor}{RGB}{234, 88, 12}     
\definecolor{Acolor}{RGB}{245, 158, 11}    
\definecolor{Scolor}{RGB}{34, 197, 94}     
\definecolor{Pcolor}{RGB}{59, 130, 246}    

\begin{document}

\title{%
    \textbf{%
        \textcolor{Gcolor}{G}%
        \textcolor{Rcolor}{R}%
        \textcolor{Acolor}{A}%
        \textcolor{Scolor}{S}%
        \textcolor{Pcolor}{P}%
    }:
    \textcolor{Gcolor}{G}eospatial
    pixel
    \textcolor{Rcolor}{R}easoning
    vi\textcolor{Acolor}{A}
    \textcolor{Scolor}{S}tructured
    \textcolor{Pcolor}{P}olicy
    learning
}

\date{}

\author{
  Chengjie Jiang\textsuperscript{1}, 
  Yunqi Zhou\textsuperscript{2}, 
  Jiafeng Yan\textsuperscript{2}, 
  Jing Li\textsuperscript{3,}\thanks{Corresponding author},
  Jiayang Li\textsuperscript{4},
  Yue Zhou\textsuperscript{3},
  Hongjie He\textsuperscript{5}
  Jonathan Li\textsuperscript{3,5}
}

\address{
  \textsuperscript{1 }Shenzhen International Graduate School, Tsinghua University, Shenzhen, China\\
  \textsuperscript{2 }School of Information, Central University of Finance and Economics, Beijing, China\\
  \textsuperscript{3 }Key Laboratory of Geographic Information Science (Ministry of Education),\\
East China Normal University, Shanghai, China\\
  \textsuperscript{4 }School of Electronic and Computer Engineering, Peking University, Shenzhen, China\\
  \textsuperscript{5 }Department of Geography and Environmental Management, University of Waterloo, Waterloo, Canada\\
}

\abstract{
Geospatial pixel reasoning aims to generate segmentation masks in remote sensing imagery directly from natural-language instructions. Most existing approaches follow a paradigm that fine-tunes multimodal large language models under supervision with dense pixel-level masks as ground truth. While effective within the training data distribution, this design suffers from two main drawbacks: \textit{(1) the high cost of large-scale dense mask annotation, and (2) the limited generalization capability of supervised fine-tuning in out-of-domain scenarios}. To address these issues, we propose \textcolor{Gcolor}{G}\textcolor{Rcolor}{R}\textcolor{Acolor}{A}\textcolor{Scolor}{S}\textcolor{Pcolor}{P}, a structured policy-learning framework that integrates a multimodal large language model with a pretrained segmentation model in a cascaded manner. To enhance generalization, we introduce \textit{PRIME}, a training paradigm that replaces supervised fine-tuning with reinforcement learning to better align reasoning and grounding behaviors with task objectives. To reduce annotation costs, we design \textit{BoP-Rewards}, which substitutes dense mask labels with bounding box and positive points. It further verifies outputs through two complementary signals: format, which constrains the reasoning and grounding structure to remain syntactically parsable, and accuracy, which evaluates the quality of predicted boxes and points. For evaluation, we train our method and all baselines on EarthReason and GeoPixInstruct, constructing an in-domain benchmark by merging their test sets. We further release \textcolor{Gcolor}{G}\textcolor{Rcolor}{R}\textcolor{Acolor}{A}\textcolor{Scolor}{S}\textcolor{Pcolor}{P}-1k, a fully out-of-domain benchmark with reasoning-intensive queries, reasoning traces, and fine-grained masks. Experimental results demonstrate state-of-the-art (SOTA) in-domain performance and up to 54\% improvement in out-of-domain scenarios, confirming that reinforcement learning with cost-aware rewards provides a robust and scalable paradigm for geospatial pixel reasoning. All code and datasets will be released publicly.
}


\maketitle

\section{Introduction}\label{sec:Introductions}

\sloppy

Remote sensing plays a crucial role in geospatial information analysis, supporting applications such as urban planning, environmental monitoring, and disaster relief~\cite{rs_review}. Traditional computer vision techniques—such as semantic segmentation—have long assisted practitioners in localizing and identifying objects within overhead imagery~\cite{classic_rs_image_classification,classic_rs_image_segmentation}.

Compared with natural images, remote sensing imagery is captured from remote sensing platforms with elevated or orbital perspectives and typically contains substantial background clutter, atmospheric distortions, extreme scale variation, and a low proportion of salient foreground objects. These characteristics complicate visual grounding and limit the transferability of methods developed for natural images, which are usually tailored to salient objects and moderate scale variation. To bridge this gap, several remote sensing-specific semantic segmentation models have been proposed to better address the unique challenges of remote sensing imagery~\cite{rsvg,geoground,isprs_seg}. 

In addition, early segmentation formulations in both natural and remote sensing domains were designed to localize all objects of a given category within a scene. However, realistic geospatial scenarios often involve diverse and complex objects of interest that defy such simple categorization. Recent approaches~\cite{rmsin,lgce,danet,rsrefseg,rsrefseg2} therefore introduce referring image segmentation into remote sensing tasks, enabling models to handle descriptive labels with positional and attribute information. Consequently, models can precisely localize specific targets instead of segmenting all similar objects in a scene. Nevertheless, such methods still require experts to manually extract and format descriptions according to predefined schemas.

\begin{figure*}[htbp]
  \centering
  \includegraphics[width=\textwidth]{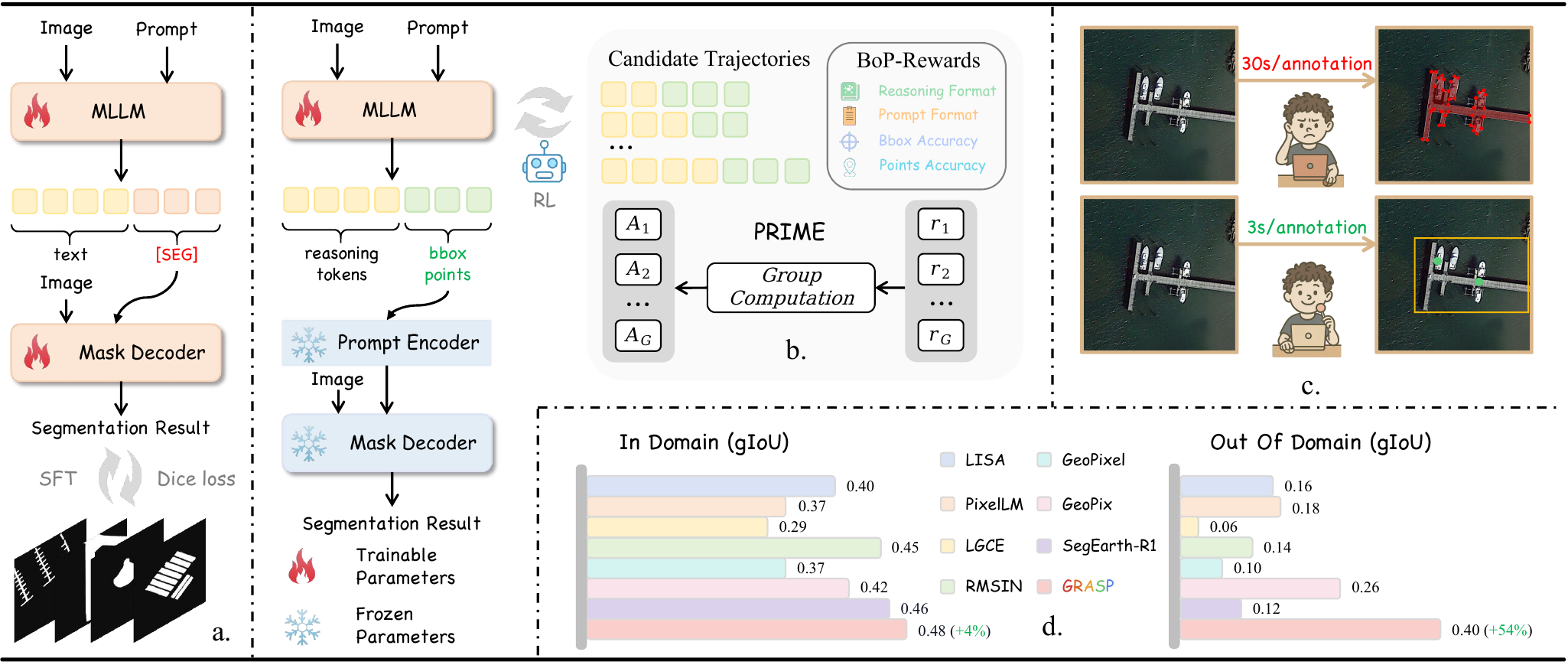}
    \caption{
    Overview of our method’s advantages. 
    (a.) Existing geospatial pixel reasoning paradigm: MLLM is trained with dense mask supervision under SFT, which leads to costly annotation and poor OOD generalization. 
    (b.) Our paradigm: \textcolor{Gcolor}{G}\textcolor{Rcolor}{R}\textcolor{Acolor}{A}\textcolor{Scolor}{S}\textcolor{Pcolor}{P} introduces \textit{PRIME}, where the MLLM is optimized purely with RL. To replace dense mask labels, we design \textit{BoP-Rewards}, which guide training through structured format and accuracy signals. 
    (c.) Annotation efficiency: fine-grained mask labeling takes $\sim$30s per sample and, for example, outlining the pier in the figure requires a polygon with 38 vertices. In contrast, our scheme only needs one bounding box and two positive points, requiring $\sim$3s. 
    (d.) Quantitative comparisons: our method achieves SOTA performance with clear gains in both in-domain (+4\%) and out-of-domain (+54\%) benchmarks.
}

 \label{fig:intro}
\end{figure*}

To further enhance model intelligence, SegEarth-R1~\cite{segearth} formulates geospatial pixel-level reasoning, where models generate masks directly from natural-language instructions. Following frameworks such as LISA~\cite{lisa,lisa++} and PixelLM~\cite{pixellm}, SegEarth-R1 aligns visual and textual embeddings, which are jointly fed into a Large Language Model (LLM) to autoregressively produce a \texttt{<SEG>} token. This token is then decoded by a mask decoder to obtain the final segmentation output. GeoPix~\cite{geopix} and GeoPixel~\cite{geopixel} concurrently adopted similar frameworks, as illustrated in Fig.~\ref{fig:intro}a. In practice, existing methods jointly train a multimodal large language model (MLLM) and a segmentation decoder under supervised fine-tuning (SFT), which relies heavily on dense mask annotations.

However, this paradigm faces two notable challenges. First, \textbf{costly supervision}: producing high-quality remote sensing segmentation annotations requires domain experts to delineate fine-grained polygons at sub-meter resolution, with careful treatment of occlusions, boundary ambiguities, and class overlaps. As illustrated in Fig.~\ref{fig:intro}c, annotating a single dock class may involve up to 38 vertices to accurately depict its polygon, highlighting the complexity of mask-level labeling. This process is highly labor-intensive and time-consuming, making large-scale mask curation prohibitively expensive. Second, \textbf{limited generalization under SFT}: SFT inherently fits the training set distribution, which often leads to training instability and overfitting. As a result, model performance substantially degrades when evaluated on out-of-domain (OOD) data. These limitations motivate two guiding questions: \textit{(\textbf{Q1}) Can complex segmentation capabilities be learned from cheaper supervision signals to reduce annotation cost? (\textbf{Q2}) Can we employ a more robust training mechanism that improves generalization, especially under OOD scenarios?}

To address \textit{\textbf{Q1}}, we reconsider the nature of the supervision signal itself. As illustrated in Fig.~\ref{fig:intro}c, annotating fine-grained segmentation masks requires delineating detailed polygons with dozens of vertices, which is prohibitively slow. This motivates us to explore coarser but still informative spatial cues—bounding boxes and positive points—as alternatives to dense masks. A preliminary user study further confirms this intuition, showing that box-and-point annotation is about 9$\times$ faster than dense mask labeling. Building on this efficiency advantage, we convert the coarse annotations into an effective training signal through \textit{BoP-Rewards} (\textit{Bo}x and \textit{P}oint \textit{Reward} \textit{S}cheme). The model is required to generate a structured reasoning trace together with spatial prompts, and its outputs are assessed by two complementary criteria: a format term, which enforces parsable reasoning and grounding structures for the downstream segmenter, and an accuracy term, which measures the quality of the predicted boxes and points. In this way, \textit{BoP-Rewards} preserve the essentials of fine-grained localization while dramatically reducing annotation cost.

Changing the supervision signal alone, however, does not resolve the generalization limitation in \textit{\textbf{Q2}}. To address this, we shift the optimization paradigm from SFT to reinforcement learning (RL). Unlike SFT, which passively fits training data distributions, RL improves the policy through interaction with reward signals and is therefore more robust to distribution shifts. Building on this insight, we adopt a \textbf{pure} RL paradigm for model optimization, termed \textit{PRIME} (\textit{P}ure \textit{R}einforcement for \textit{I}nstruction-to-\textit{M}ask \textit{E}xecution). Since the pretrained MLLM and segmentation model already encode strong priors for reasoning and fine-grained segmentation, \textit{PRIME} avoids retraining them from scratch and instead focusing on refining their decision process through carefully designed rewards. Concretely, the segmentation model is kept frozen, while the MLLM is optimized with Grouped Relative Policy Optimization (GRPO)~\cite{deepseekmath} to maximize the \textit{BoP-Rewards}. 

By coupling scalable, mask-free guidance with reward-driven policy optimization, we arrive at \textcolor{Gcolor}{G}\textcolor{Rcolor}{R}\textcolor{Acolor}{A}\textcolor{Scolor}{S}\textcolor{Pcolor}{P}, the full pipeline illustrated in Fig.~\ref{fig:intro}b. Given a vision–language instruction, the MLLM generates a structured reasoning trace and task-relevant spatial prompts (bounding boxes and positive points), which a pretrained segmentation model then converts into the final mask.

For fair comparison, we train our method and all baselines on the training sets of two representative datasets: the geospatial pixel reasoning dataset EarthReason\cite{segearth} and the remote-sensing referring image segmentation dataset GeoPixInstruct\cite{geopix}. We construct a single in-domain benchmark by merging their official test sets and evaluate all models under this unified protocol. Beyond the in-domain setting, we curate \textcolor{Gcolor}{G}\textcolor{Rcolor}{R}\textcolor{Acolor}{A}\textcolor{Scolor}{S}\textcolor{Pcolor}{P}-1k, a finely annotated benchmark specifically designed to test OOD generalization. It is built from a high-quality image pool entirely disjoint from the training data, and we employ a semi-automated pipeline to generate reasoning-intensive queries, reasoning traces, and fine-grained segmentation annotations. As shown in Fig.~\ref{fig:intro}d, our method achieves significant gains under both in-domain and OOD settings, highlighting its robustness to distribution shifts.

The main contributions can be summarized as follows:

\begin{itemize}
  \item To reduce the prohibitive cost of dense pixel annotations, we propose \textit{BoP-Rewards}, a mask-free reward scheme that replaces fine-grained masks with bounding boxes and positive points, and transforms them into structured policy signals via format and accuracy terms. This design enables the model to learn precise localization behaviors at a fraction of the annotation cost. 
  
  \item To address the limited generalization caused by SFT, we introduce \textit{PRIME}, an SFT-free training paradigm that optimizes only the MLLM with GRPO. By leveraging and stimulating the pretrained models’ inherent reasoning and segmentation capabilities, \textit{PRIME} achieves substantially stronger robustness to distribution shifts. 
  
  \item To rigorously evaluate OOD performance, we release \textcolor{Gcolor}{G}\textcolor{Rcolor}{R}\textcolor{Acolor}{A}\textcolor{Scolor}{S}\textcolor{Pcolor}{P}-1k, a benchmark with high-quality imagery, reasoning-intensive queries, structured reasoning traces, and fine-grained masks. Experiments under a fair evaluation protocol show that our method not only achieves SOTA results in-domain but also delivers up to 54\% improvements on OOD benchmarks.
\end{itemize}

\begin{figure*}[t!]
  \centering
  \includegraphics[height=0.48\textheight,width=\textwidth]{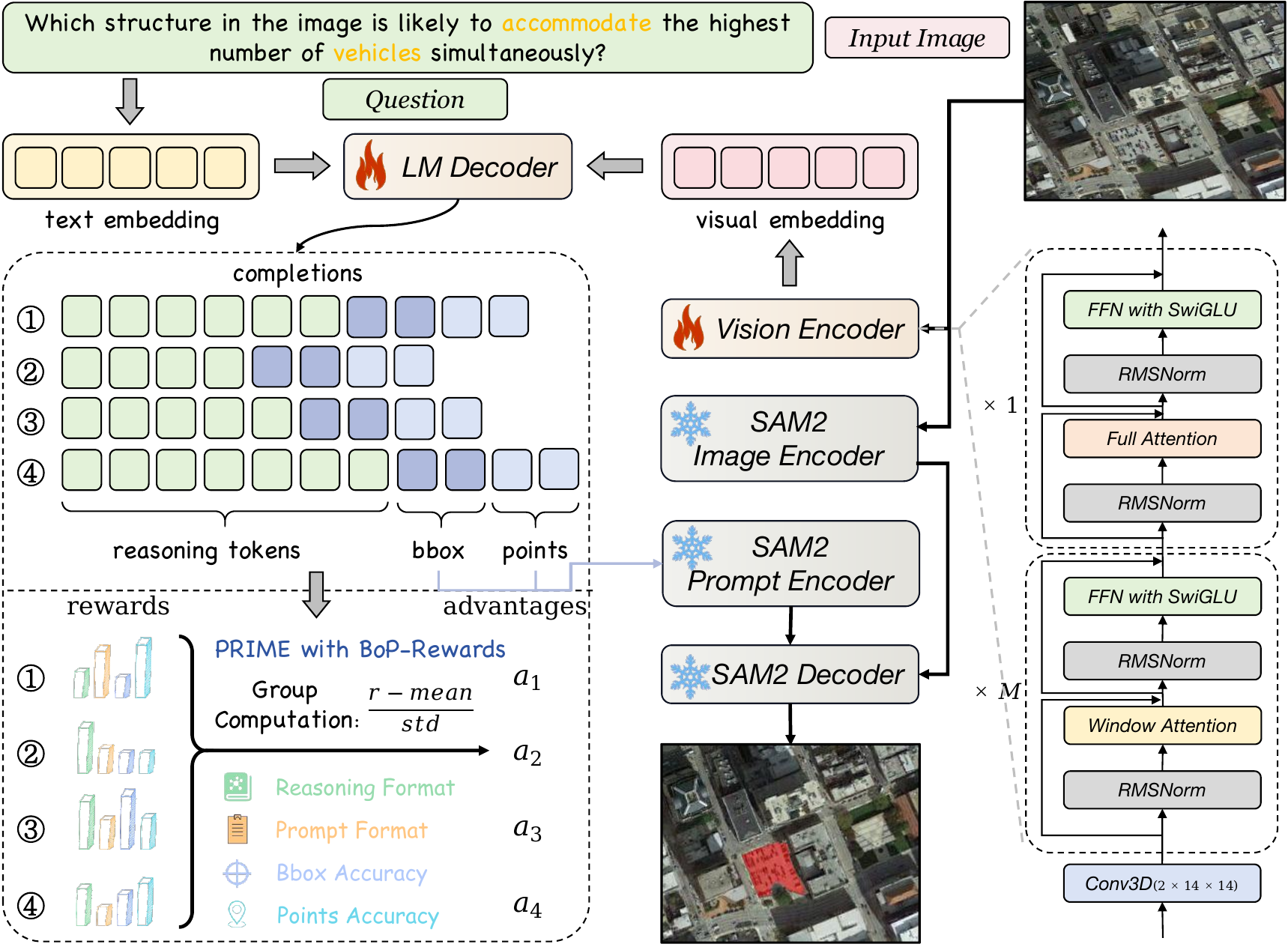}
  \caption{\textbf{Overview of the model architecture and the training workflow.}
The framework consists of two main components: an MLLM and a segmentation model. The MLLM comprises a vision encoder and a language model (LM) decoder. The LM decoder output contains both reasoning tokens and spatial grounding predictions in the form of bounding boxes and positive points. The spatial grounding predictions are fed into the SAM2 prompt encoder as prompts, while the original image is simultaneously input to the SAM2 image encoder. Finally, the SAM2 decoder combines both sources of information to produce the segmentation mask.}
  \label{fig:data construction pipeline}
\end{figure*}

\section{Related Works}\label{sec:Related works}
Segmentation tasks in remote sensing have evolved from semantic segmentation to referring image segmentation, and more recently to geospatial pixel reasoning. 

\subsection{Semantic Segmentation in Remote Sensing}
Semantic segmentation is a fundamental task in remote sensing, aiming to assign semantic labels to each pixel in aerial or satellite imagery for applications such as land cover mapping and urban monitoring~\cite{remote_sensing_seg_review}. With the rise of deep learning, CNN-based methods such as FCN~\cite{FCN}, SegNet~\cite{segearth}, U-Net~\cite{unet}, and DeepLab~\cite{deeplab} became widely adopted, demonstrating strong performance through encoder–decoder architectures and multi-scale feature aggregation. More recently, transformer-based models have been introduced to address the limitations of CNNs in modeling long-range dependencies. Representative examples include ViT~\cite{vit}, Swin Transformer~\cite{swin}, and SegFormer~\cite{segformer}, alongside remote sensing–specific variants such as Efficient Transformer~\cite{efficient_transformer}, STransFuse~\cite{stransfuse}, and UNetFormer~\cite{unetformer}. These approaches have improved robustness in complex remote sensing scenarios. Nevertheless, semantic segmentation remains constrained by its reliance on predefined categories. As a result, such models cannot flexibly handle diverse or fine-grained targets that do not align with the annotation schema. Such limitations highlight the need for more interactive segmentation paradigms that can adapt to complex and varied user demands.

\subsection{Referring Image Segmentation in Remote Sensing}
Referring image segmentation (RIS) generalizes semantic segmentation by replacing predefined category labels with natural-language queries that incorporate attributes or spatial relations. This formulation enables more flexible and fine-grained object localization, which has motivated its recent adaptation to remote sensing scenarios. Yuan et al. introduced the RefSegRS dataset together with the language-guided cross-scale enhancement (LGCE) module~\cite{lgce}, marking the first Transformer-based approach for referring image segmentation in aerial imagery. LGCE leverages linguistic features to guide multi-scale visual enhancement and integrates deep semantic with shallow spatial features. This design improves the segmentation of small and spatially dispersed objects that frequently occur in remote sensing imagery. Building upon this work, Liu et al. proposed Rotated Multi-Scale Interaction Network (RMSIN)~\cite{rmsin}, which explicitly addresses the challenges of multi-scale variation and rotated objects. RMSIN incorporates an Intra-scale Interaction Module to refine features within each scale, a Cross-scale Interaction Module to integrate information across scales, and an Adaptive Rotated Convolution to enhance robustness to diverse orientations. To strengthen evaluation, RMSIN also curates the RRSIS-D dataset containing over 17k image–caption–mask triplets, providing a more comprehensive benchmark for remote sensing referring image segmentation. 

More recently, Dual Alignment Network (DANet)~\cite{danet} revisited the dominant paradigm of “implicit optimization,” where vision–language fusion and segmentation are only loosely aligned. It introduces a dual-alignment strategy consisting of explicit affinity alignment to mitigate the domain gap between modalities, and a reliable agent alignment module to reinforce semantic awareness while suppressing background noise. These designs collectively improve robustness and generalization across complex scenarios. Despite these advances, remote sensing referring image segmentation still remains constrained. Current models often rely on relatively template-based textual inputs, which limits their adaptability to natural language variability. Moreover, they lack higher-level reasoning ability, reducing effectiveness when confronted with more complex or reasoning-intensive instructions.

\subsection{Geospatial Pixel Reasoning}

To address the limitations of traditional Referring Image Segmentation (RIS) in remote sensing, SegEarth-R1~\cite{segearth} introduced the task of \textit{geospatial pixel reasoning}, which goes beyond explicit referring expressions to support open-ended natural language queries. These queries often reflect real-world needs, such as “identifying regions with water resources” or “locating safe areas for large gatherings,” and require models to reason over spatial context and domain knowledge.  

Most recent methods tackling this task adopt the LISA-style framework~\cite{lisa,lisa++}, where an MLLM generates a task-specific \texttt{[SEG]} token, which is passed to a pretrained segmentation backbone like SAM~\cite{sam,sam2} to generate the final mask. This design has become a standard foundation for geospatial pixel reasoning.

Based on this framework, SegEarth-R1 applies a hierarchical Swin Transformer encoder with aggressive token compression to handle ultra-high-resolution inputs. It is trained on the EarthReason benchmark, which includes implicit queries paired with fine-grained masks. By incorporating architectural adaptations and domain-specific queries, SegEarth-R1 shifts the focus from direct instruction following to implicit reasoning in complex scenes.

GeoPix~\cite{geopix} also builds on the LISA paradigm but focuses on improving segmentation consistency. It introduces a class-wise learnable memory (CLM) module that captures shared geo-context across categories, which helps reduce confusion in scenes with dense or overlapping objects. Additionally, GeoPix constructs the large-scale GeoPixInstruct dataset with text, bounding boxes, and pixel-level masks, and uses a two-stage training strategy that first aligns text–image pairs and then fine-tunes segmentation. Compared to SegEarth-R1, which emphasizes reasoning, GeoPix aims to improve stability and scalability.

GeoPixel~\cite{geopixel}, in contrast, focuses on adapting the LISA framework to interactive, high-resolution settings. It introduces a dynamic partitioning strategy that splits input images into global and local views, allowing processing of up to 4K resolution. Instead of memory modules, it uses a frozen SAM-2 encoder and a lightweight pixel decoder to directly map \texttt{[SEG]} tokens to segmentation masks. This design improves scalability and supports fine-grained visual grounding in dialogue-style interaction. While SegEarth-R1 and GeoPix emphasize reasoning and memory consistency respectively, GeoPixel emphasizes resolution handling and efficient deployment.

Despite these advances, current methods still rely heavily on dense mask annotations and are prone to domain overfitting, which limits generalization. Meanwhile, reinforcement learning (RL) has shown strong reasoning capabilities~\cite{deepseekmath,segzero}, making it a promising direction for improving instruction-based segmentation in remote sensing.

\section{Method}\label{sec:Method}

This section details the proposed \textcolor{Gcolor}{G}\textcolor{Rcolor}{R}\textcolor{Acolor}{A}\textcolor{Scolor}{S}\textcolor{Pcolor}{P} framework, beginning with a pipeline overview (Sec.~\ref{subsec:pipeline}), followed by the model architecture and PRIME training paradigm (Sec.~\ref{subsec:model}), and concluding with the cost-aware BoP-Rewards scheme (Sec.~\ref{subsec:reward}).

\begin{figure*}[!t]
  \centering
  \includegraphics[width=\textwidth]{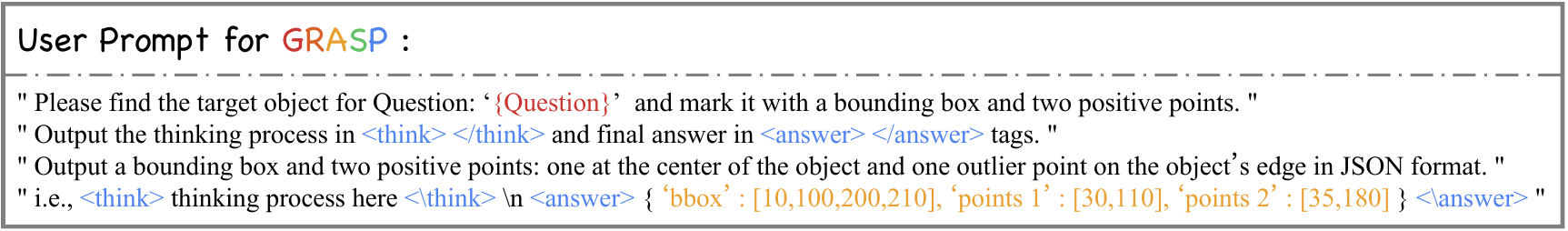}
  \caption{\textbf{User prompt for \textcolor{Gcolor}{G}\textcolor{Rcolor}{R}\textcolor{Acolor}{A}\textcolor{Scolor}{S}\textcolor{Pcolor}{P}.} \textcolor{Gcolor}{\{Question\}} is replaced with geospatial pixel reasoning question $Q$ in both training and inference stage.}
  \label{fig:prompt}
\end{figure*}



\subsection{Pipeline Formulation}\label{subsec:pipeline}

Given a remote sensing image $I \in \mathbb{R}^{H \times W \times 3}$ and a reasoning-intensive query $Q$, the objective of geospatial pixel reasoning is to predict a segmentation mask $M \in \{0,1\}^{H \times W}$ whose covered region precisely corresponds to the region in $I$ that answers $Q$. Formally, the task can be expressed as learning a mapping:
\begin{equation}
    f: (I, Q) \;\mapsto\; M.
\end{equation}

Directly learning $f$ under dense mask supervision is both annotation-intensive and prone to overfitting. To address this, we decouple the reasoning and segmentation processes into two complementary modules. Following this design, we optimize an MLLM $\pi_\theta$ under \textit{PRIME} training paradigm. 

Conditioned on $(I, Q)$, the MLLM produces a structured reasoning trace $\mathcal{R}$ together with spatial grounding signals:
\begin{equation}
    (\mathcal{R}, B, P_1, P_2) \;=\; \pi_\theta(I, Q),
\end{equation}
where $\mathcal{R}$ is a token sequence encoding the stepwise reasoning process, $B \in \mathbb{R}^4$ denotes a bounding box, and $P_1, P_2 \in \mathbb{R}^2$ are positive points located within the target region. These cues provide a coarse yet informative localization prior.

The second stage employs a pretrained segmentation model $\phi$ that accepts $(I, B, P_1, P_2)$ as prompts and outputs the final dense mask:
\begin{equation}
    M \;=\; \phi\!\left(I, B, P_1, P_2\right).
\end{equation}

Overall, the cascaded pipeline can thus be summarized as follows:
\begin{equation}
    f(I, Q) \;=\; \phi\!\big(I,\, \pi_\theta(I, Q)\big),
\end{equation}
where the MLLM specializes in reasoning and spatial grounding, while the segmentation model ensures fine-grained pixel delineation. This division leverages the complementary strengths of both modules: the MLLM provides instruction-driven reasoning and localization, while the pretrained segmenter translates coarse spatial cues into precise masks.

\subsection{\textcolor{Gcolor}{G}\textcolor{Rcolor}{R}\textcolor{Acolor}{A}\textcolor{Scolor}{S}\textcolor{Pcolor}{P} Model}\label{subsec:model}

The geospatial pixel reasoning task requires both strong vision–language reasoning and fine-grained segmentation. Recent advances in MLLMs~\cite{llava,llavanext,internvl,qwenvl,qwenvl25,deepseekvl} have demonstrated powerful reasoning capabilities, while pretrained segmentation models~\cite{sam,sam2} offer high-precision mask generation. Motivated by these complementary strengths, we design a cascaded framework that integrates an MLLM with a segmentation model. To further enhance generalization, we replace conventional SFT with an RL paradigm, termed \textit{PRIME}, which optimizes only the MLLM while keeping the segmentation model frozen. Fig.~\ref{fig:data construction pipeline} provides an overview of the proposed architecture, denoted as \textcolor{Gcolor}{G}\textcolor{Rcolor}{R}\textcolor{Acolor}{A}\textcolor{Scolor}{S}\textcolor{Pcolor}{P}.

\paragraph{Reasoning via Qwen.}
For the reasoning stage, we adopt Qwen2.5-VL~\cite{qwenvl25}, a powerful MLLM capable of visual grounding with bounding boxes, but lacking native fine-grained segmentation capability. To adapt it to the geospatial pixel reasoning task, we set the parameters of its vision encoder to be trainable, so that the model can better accommodate the unique characteristics of remote sensing imagery. We also make the LM decoder trainable, since it must be guided to follow the task-specific output schema required in our pipeline. In particular, the decoder is trained to produce a structured reasoning trace $\mathcal{R}$ along with spatial prompts, including a bounding box $B$ and two positive points $P_1, P_2$. This enables the model to go beyond generic grounding and explicitly provide the structured signals required by our \textit{BoP-Rewards}.  

Formally, the reasoning stage can be expressed as:
\begin{equation}
    (\mathcal{R}, B, P_1, P_2) \;=\; \mathrm{MLLM}(I, Q;\, \theta_{\text{vis}}, \theta_{\text{dec}}),
\end{equation}
where $I$ and $Q$ denote the input image and query, while $\theta_{\text{vis}}$ and $\theta_{\text{dec}}$ are the trainable parameters of the vision encoder and LM decoder, respectively.

\paragraph{Segmentation via SAM.}
For the segmentation stage, we employ SAM2~\cite{sam2}, a SOTA open-world segmentation model. SAM2 accepts diverse prompts, including dense prompts such as masks and sparse prompts such as bounding boxes and points. To reduce annotation cost and align with our \textit{BoP-Rewards} design, we deliberately use only sparse prompts—specifically the bounding box and positive points predicted by the MLLM. These cues are embedded by SAM2’s prompt encoder, and together with the image embedding from SAM2’s image encoder, are fed to the SAM2's mask decoder to generate the final fine-grained segmentation mask $M$:
\begin{equation}
    M \;=\; \mathrm{SAM2}\big(\mathrm{Enc}_{\text{img}}(I),\, \mathrm{Enc}_{\text{prompt}}(B, P_1, P_2)\big).
\end{equation}
During training, all SAM2 parameters are kept frozen to preserve its pretrained segmentation priors and to stabilize optimization.

\paragraph{\textit{PRIME} Training Paradigm.}
In the \textit{PRIME} paradigm, the MLLM is the only component updated during training, while the pretrained segmentation model remains frozen. Training is based on the principle of GRPO~\cite{deepseekmath}, which is adapted in our setting as the \textit{PRIME} objective. For each query $q$ sampled from $P(Q)$, the policy model $\pi_{\theta_{\mathrm{old}}}$ generates a group of $G$ candidate reasoning traces and spatial prompts $\{o_1, o_2, \dots, o_G\}$. Each candidate receives a scalar reward $\{r_1, r_2, \dots, r_G\}$ computed from the \textit{BoP-Rewards} in Sec.~\ref{subsec:reward}, which jointly captures format compliance and grounding accuracy.

To reduce sensitivity to absolute reward magnitudes, rewards are standardized into relative advantages:
\begin{equation}
    A_i = \frac{r_i - \mathrm{mean}(\{r_1,\dots,r_G\})}{\mathrm{std}(\{r_1,\dots,r_G\})}.
\end{equation}

The \textit{PRIME} objective is then constructed from three parts. First, the policy ratio $\frac{\pi_{\theta}(o_i \mid q)}{\pi_{\theta_{\mathrm{old}}}(o_i \mid q)}$ compares the likelihood of the new policy producing candidate $o_i$ relative to the old policy. Multiplying this ratio by $A_i$ encourages the model to increase the probability of above-average candidates, such as accurate boxes or well-structured outputs. Second, clipping the ratio within $[1-\varepsilon, 1+\varepsilon]$ prevents unstable updates, ensuring steady improvement even when reward variance is large. Third, a KL divergence term $\mathrm{D}_{\mathrm{KL}}\!\left(\pi_{\theta} \,\|\, \pi_{\mathrm{ref}}\right)$ anchors the policy to a frozen reference model $\pi_{\mathrm{ref}}$ (pretrained Qwen2.5-VL in our case). This constraint prevents RL from eroding the pretrained reasoning and bounding-box grounding abilities, while still allowing refinement of new behaviors such as point prediction and structured reasoning.

In summary, these elements form the final \textit{PRIME} objective:

\begin{strip}
\begin{equation}
\mathcal{J}_{\mathrm{PRIME}}(\theta) =
\mathbb{E}_{q, \{o_i\}}
\left[
\frac{1}{G} \sum_{i=1}^{G}
\min\!\left(
\frac{\pi_{\theta}(o_i \mid q)}{\pi_{\theta_{\mathrm{old}}}(o_i \mid q)} A_i,\;
\mathrm{clip}\!\left(\frac{\pi_{\theta}(o_i \mid q)}{\pi_{\theta_{\mathrm{old}}}(o_i \mid q)}, 1-\varepsilon, 1+\varepsilon\right) A_i
\right)
-
\beta\,\mathrm{D}_{\mathrm{KL}}\!\left(\pi_{\theta} \,\|\, \pi_{\mathrm{ref}}\right)
\right],
\label{eq:prime}
\end{equation}
\end{strip}
where $\varepsilon$ and $\beta$ are hyperparameters. 

\subsection{BoP-Rewards}\label{subsec:reward}

\textit{BoP-Rewards} provide a rule-based reward scheme tailored for geospatial pixel reasoning. We define \textbf{five} rewards organized into two dimensions: 1) \textbf{format compliance}, which ensures that the model output follows a standardized structure and can be parsed into reasoning trace, bounding box, and points; and 2) \textbf{localization accuracy}, which evaluates the precision of the predicted bounding box and points relative to GT. The total training signal is obtained by summing the five components, allowing reinforcement learning to guide both the structure and accuracy of the model’s outputs.

\paragraph{Notation.}
Bounding boxes are axis-aligned. Let the ground-truth (GT) box be $B_g=(x_1^g,y_1^g,x_2^g,y_2^g)$ with width $w_g = x_2^g - x_1^g$ and height $h_g = y_2^g - y_1^g$, and define $s_{\min} = \min(w_g, h_g)$, $s_{\max} = \max(w_g, h_g)$. 
The predicted bounding box is denoted $B_p=(x_1^p,y_1^p,x_2^p,y_2^p)$, and the corresponding center coordinates and dimensions are $(c_x^p, c_y^p, w^p, h^p)$.  
Predicted points are denoted $p_1^p, p_2^p$ and GT points are $p_1^g, p_2^g$.

\subsubsection*{Format Rewards}

\paragraph{(1) Reasoning Format Reward.}
This discrete reward enforces the model to output a structured reasoning process. 
The output must include a reasoning chain enclosed within \texttt{<think>} and \texttt{</think>} tags, and a grounding answer enclosed within \texttt{<answer>} and \texttt{</answer>} tags. 
If the output fully meets these format requirements, the reward is 1; otherwise, it is 0.

\paragraph{(2) Prompt Format Reward.}
This reward regulates the output format of the grounding answer. As illustrated in Fig.~\ref{fig:prompt}, the content inside \texttt{<answer>}\ldots\texttt{</answer>} must strictly contain a bounding box, a point~1, and a point~2 following the specified schema. The reward will be 1 only when the output exactly follows this schema; otherwise, it is 0.

\subsubsection*{Localization Rewards}

\paragraph{(3) Bbox IoU Reward.}
We calculate the intersection-over-union (IoU) between the predicted bounding box $B_p$ and the ground-truth bounding box $B_g$ as
\begin{equation}
    \mathrm{IoU}(B_p, B_g) = \frac{\mathrm{area}(B_p \cap B_g)}{\mathrm{area}(B_p \cup B_g)}.
\end{equation}
If the IoU is greater than $0.5$, the reward is 1; otherwise, it is 0.

\paragraph{(4) Bbox Distance Reward.}
Although IoU provides a standard measure of overlap, it is relatively insensitive to small positional deviations and degenerates into a binary signal when applied with a fixed threshold. To provide a smoother optimization signal and ensure fairness across different object scales, we introduce a scale-normalized distance reward.  
Let $B_p=(c_x^p, c_y^p, w^p, h^p)$ and $B_g=(c_x^g, c_y^g, w^g, h^g)$ be the center coordinates and dimensions of the predicted and GT boxes. The normalized L1 distance is computed as follows:
\begin{equation}
    d_{\text{bbox}} = \frac{1}{2}\left(
\frac{|c_x^p - c_x^g|}{w_g} + \frac{|c_y^p - c_y^g|}{h_g}
\right),
\end{equation}
where the width and height terms are omitted because their normalization is symmetric to the center offset terms. The soft reward is defined as follows:
\begin{equation}
    R_{\text{bbox-dist}} = \max\left(0,\,1 - \frac{d_{\text{bbox}}}{0.5}\right),
\end{equation}
which decays linearly with the normalized distance.

\paragraph{(5) Points Accuracy Reward.}
We first require both predicted points to lie inside the GT bounding box. 
The scale-normalized L1 distances for point~1 and point~2 are computed as follows:
\begin{equation}
    d_1 = \frac{\lVert p_1^p - p_1^g\rVert_1}{s_{\min}}, \quad
    d_2 = \frac{\lVert p_2^p - p_2^g\rVert_1}{s_{\max}}.
\end{equation}
The overall point distance is then computed as the arithmetic mean of $d_1$ and $d_2$:
\begin{equation}
    S = \frac{d_1 + d_2}{2}.
\end{equation}
A reward of $1$ is assigned if both points lie within the GT bounding box and $S < 0.5$; otherwise, the reward is set to $0$.

\begin{figure}[!t]
\centering
\includegraphics[width=0.8\columnwidth]{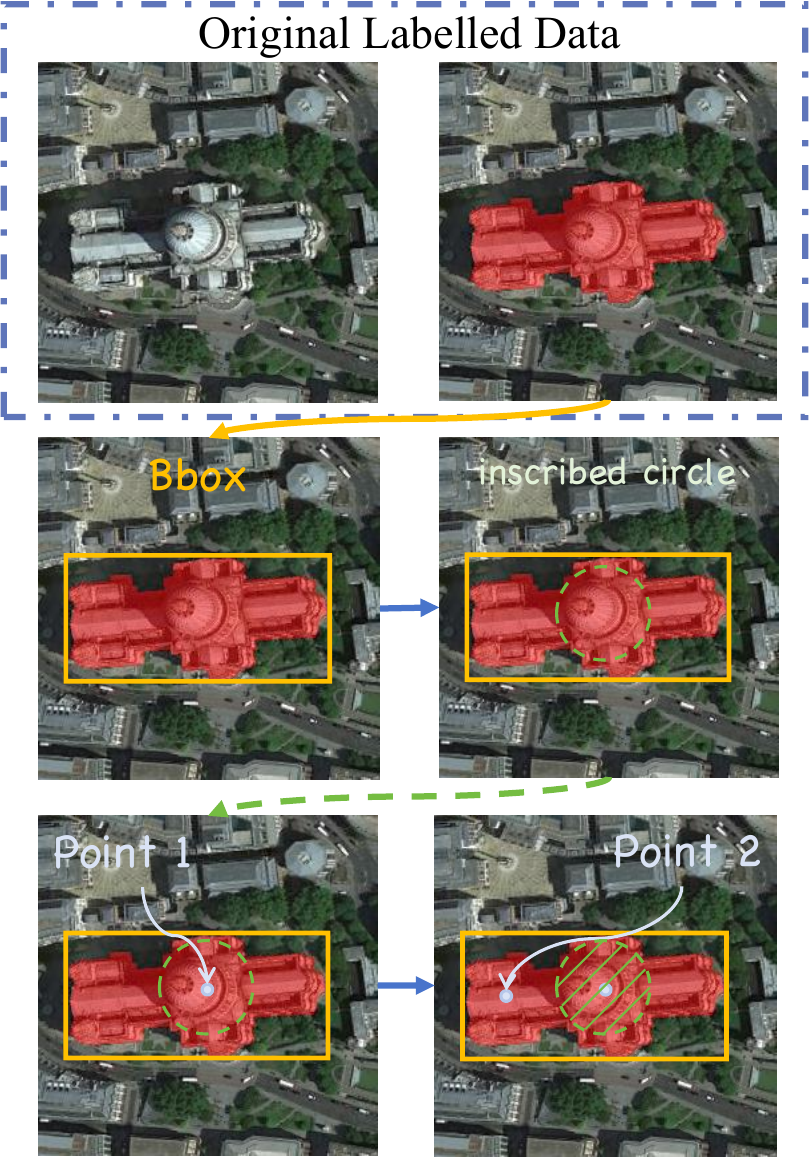}
\caption{Reconstruction pipeline for training data. We transform dense segmentation masks into sparse supervision comprising a bounding box and two positive points.}
\label{fig:reorganize}
\end{figure}

\begin{figure*}[!t]
  \centering
  \includegraphics[width=\textwidth]{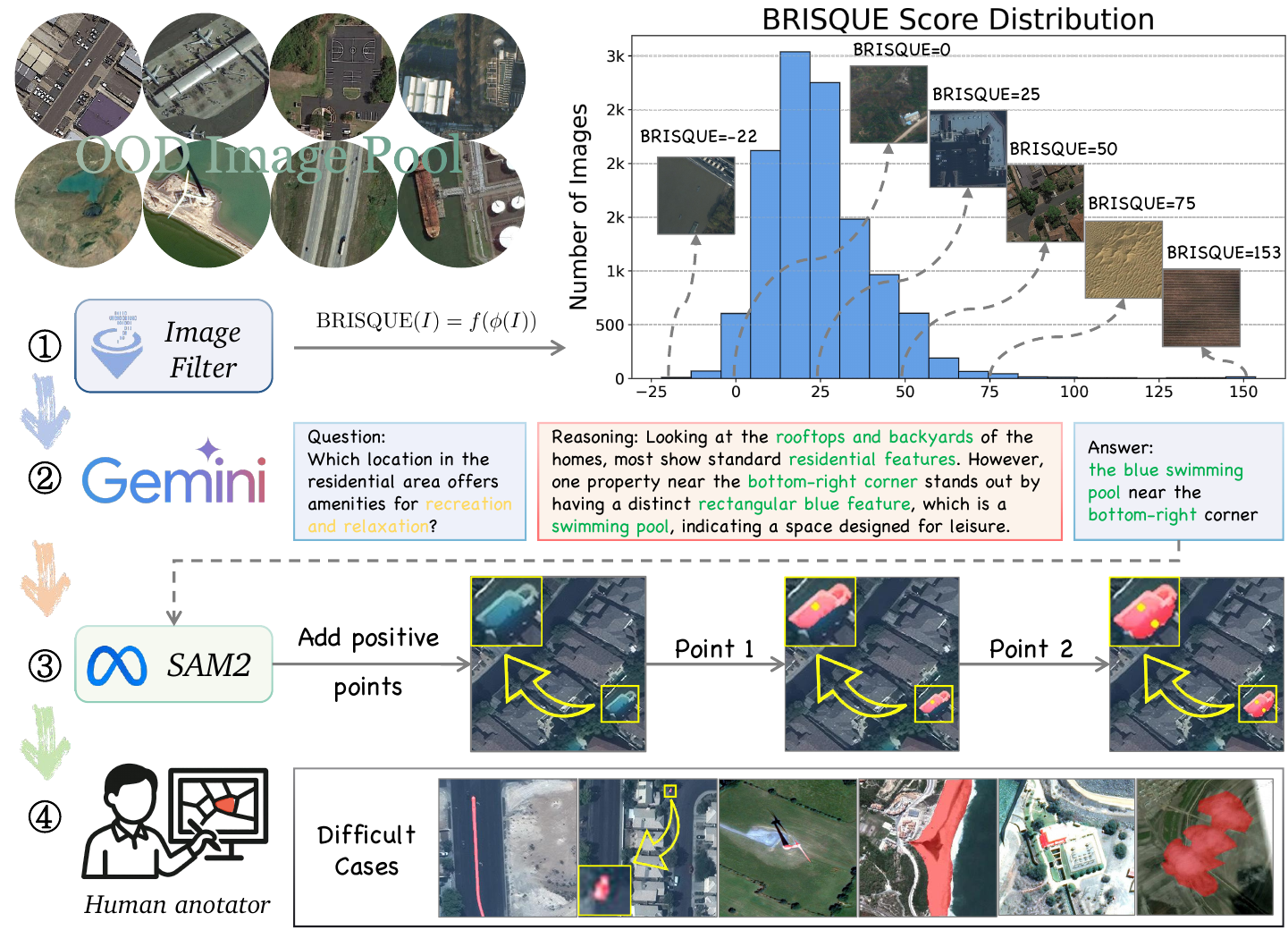}
  \caption{\textbf{Overview of the \textcolor{Gcolor}{G}\textcolor{Rcolor}{R}\textcolor{Acolor}{A}\textcolor{Scolor}{S}\textcolor{Pcolor}{P}-1k construction pipeline.} We first curate seven OOD image pools and filter out low-quality images with BRISQUE scores above 50. For the remaining images, Gemini-2.5-Pro is used to generate reasoning-intensive questions, complete with detailed explanations and spatially grounded answers. Human annotators then click on positive points indicated by the answer and leverage SAM2 for rapid segmentation. In challenging cases where SAM2 fails, manual annotations are performed using LabelMe.}
  \label{fig:grasp_bench}
\end{figure*}

\begin{figure*}[!t]
  \centering
  \includegraphics[width=\textwidth]{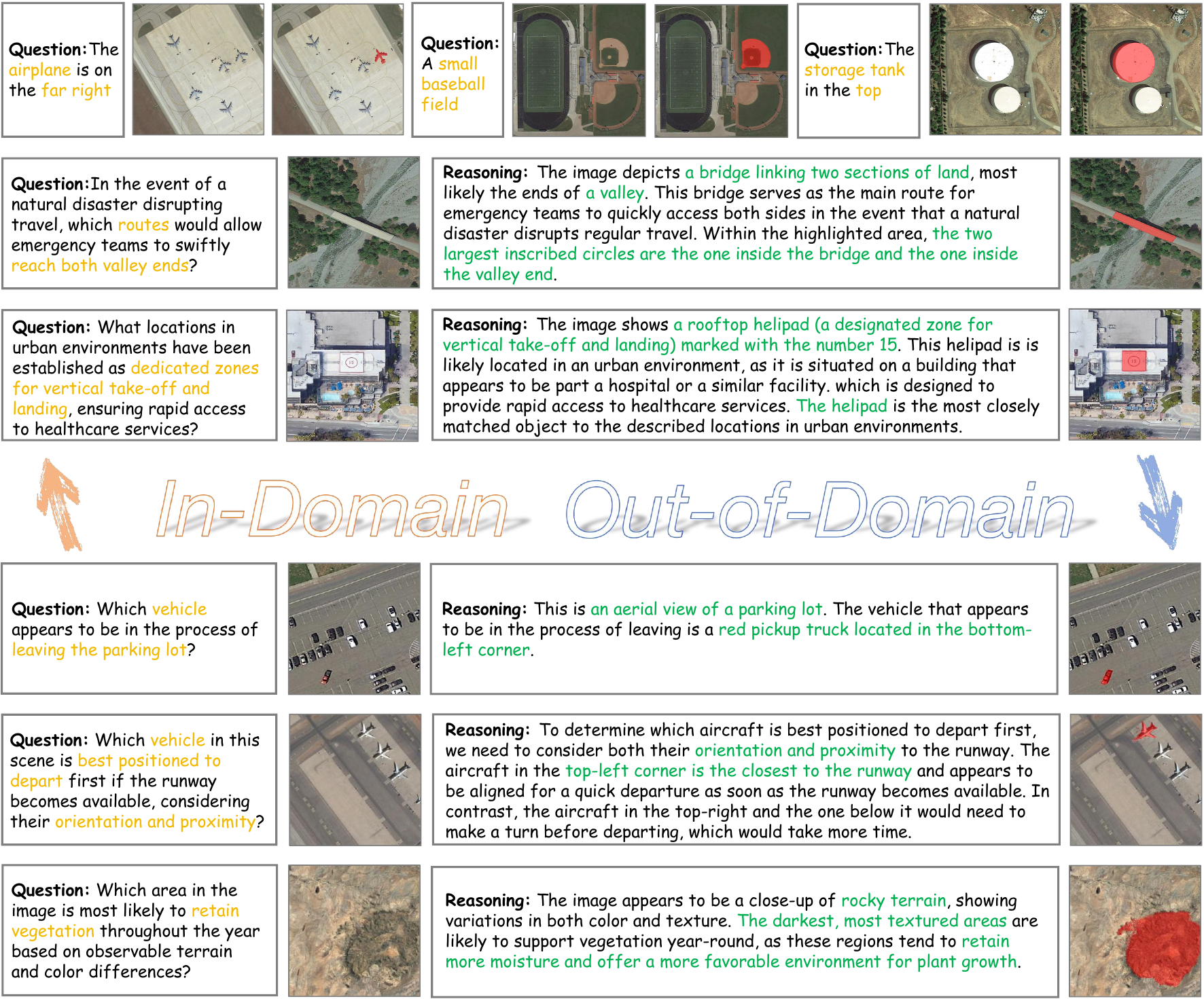}
  \caption{\textbf{Qualitative results of the \textcolor{Gcolor}{G}\textcolor{Rcolor}{R}\textcolor{Acolor}{A}\textcolor{Scolor}{S}\textcolor{Pcolor}{P} model.} The top three rows present cases from the in-domain test set, with the first row derived from GeoPixInstruct and the second and third rows from EarthReason. The bottom three rows show cases from \textcolor{Gcolor}{G}\textcolor{Rcolor}{R}\textcolor{Acolor}{A}\textcolor{Scolor}{S}\textcolor{Pcolor}{P}-1k, an OOD test set. Our model demonstrates the ability to generate both detailed reasoning chains and fine-grained segmentation masks.}
  \label{fig:good_cases}
\end{figure*}

\begin{figure*}[!t]
  \centering
  \includegraphics[width=\textwidth]{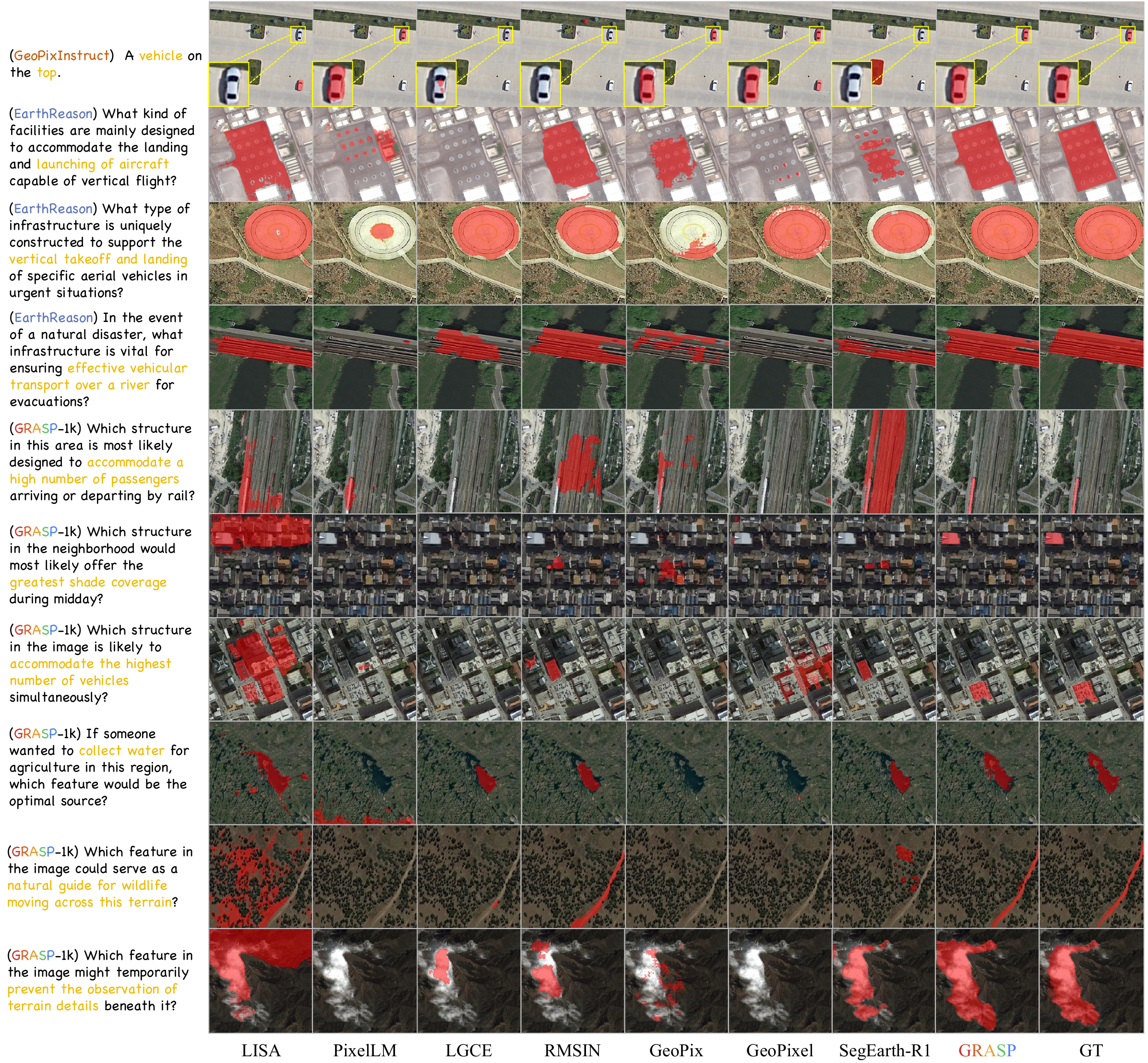}
  \caption{Qualitative comparison between our \textcolor{Gcolor}{G}\textcolor{Rcolor}{R}\textcolor{Acolor}{A}\textcolor{Scolor}{S}\textcolor{Pcolor}{P} model and seven other state-of-the-art methods on both in-domain and out-of-domain test sets. The first to fourth rows correspond to results on the in-domain test set, while the fifth to tenth rows show results on the out-of-domain test set.}
  \label{fig:qualitative comparison.}
\end{figure*}

\section{Data Construction}\label{sec:Data Construction}

This section introduces the construction of two types of data: 1) the conversion of existing dense segmentation labels into sparse signals comprising a bounding box and two positive points per data point, and 2) the construction pipeline of the \textcolor{Gcolor}{G}\textcolor{Rcolor}{R}\textcolor{Acolor}{A}\textcolor{Scolor}{S}\textcolor{Pcolor}{P}-1k benchmark, a fine-grained geospatial pixel reasoning dataset designed for rigorous evaluation under OOD conditions.

\subsection{Training Data Construction}

To ensure fair comparison with existing geospatial pixel reasoning methods, we do not introduce any external grounding datasets. Instead, we construct our training data using two representative benchmarks in this domain. The first is the EarthReason dataset~\cite{segearth}, which defines the geospatial pixel reasoning task and contains 2,371 training samples together with 1,928 test samples. The second is the GeoPixInstruct dataset~\cite{geopix}, a large-scale remote sensing referring image segmentation benchmark, which offers 115,741 training samples and 7,932 validation samples. For our experiments, we use the training set of both datasets and construct an in-domain benchmark by merging the 1,928 test samples from EarthReason with the 7,932 validation samples from GeoPixInstruct to form an in-domain benchmark. Finally, to align with our \textit{BoP-Rewards} scheme, we convert the dense segmentation masks in these datasets into bounding box and points-based supervision signals. 

As shown in Fig.~\ref{fig:reorganize}, given a binary mask $\mathcal{M}$ representing the dense object segmentation, we first extract the bounding box $\mathbf{b} = [x_\text{min}, y_\text{min}, x_\text{max}, y_\text{max}]$ by locating the extreme foreground pixels of $\mathcal{M}$. Then, to generate two positive points $\mathbf{p}_1$ and $\mathbf{p}_2$ as sparse supervision, we adopt the following strategy:  
- $\mathbf{p}_1$ is the center of the maximal inscribed circle within the mask, approximated by the point with the largest Euclidean distance to the mask boundary.  
- $\mathbf{p}_2$ is a supplementary point sampled from the outer ring of the mask beyond the inscribed circle. If no such candidate exists, we select the farthest point on the mask boundary from $\mathbf{p}_1$.

\subsection{\textcolor{Gcolor}{G}\textcolor{Rcolor}{R}\textcolor{Acolor}{A}\textcolor{Scolor}{S}\textcolor{Pcolor}{P}-1k Construction Pipeline}

To construct the \textcolor{Gcolor}{G}\textcolor{Rcolor}{R}\textcolor{Acolor}{A}\textcolor{Scolor}{S}\textcolor{Pcolor}{P}-1k benchmark, we curated images drawn entirely from OOD sources, ensuring a fully disjoint benchmark for evaluation. In contrast, the training data for our models is derived from two existing benchmarks: EarthReason, whose imagery originates from the Million-AID~\cite{million} and DIOR~\cite{dior} datasets, and GeoPixInstruct, which is built on HRSC2016~\cite{hrsc}, DOTA-V2.0~\cite{dota}, and FAIR1M-2.0~\cite{fair1m}. By deliberately excluding these sources, we ensure that \textcolor{Gcolor}{G}\textcolor{Rcolor}{R}\textcolor{Acolor}{A}\textcolor{Scolor}{S}\textcolor{Pcolor}{P}-1k serves as a fully disjoint benchmark for evaluating OOD generalization. As summarized in Table~\ref{tab:dataset_summary}, the selected datasets cover a wide spectrum of image sizes (from small patches such as $256\times256$ to ultra-large scenes exceeding $16\!K\times16\!K$) and ground resolutions (ranging from 0.08\,m/pixel to 30\,m/pixel). This diversity ensures that the benchmark captures both variations in spatial scale and differences in acquisition settings, thereby providing a rigorous test of model robustness.

\renewcommand{\arraystretch}{1.5}
\begin{table}[htbp]
  \centering
  \resizebox{\linewidth}{!}{
  \begin{tabular}{lcc}
    \toprule
    \textbf{Data Source} & \textbf{Image Size} & \textbf{Resolution} \\
    \midrule
    CVUSA~\cite{cvusa} & 800 & 0.08m \\
    NWPU-RESISC45~\cite{nwpu} & 256 & 0.2$\sim$30m \\
    CrowdAI~\cite{crowdAI} & -- & $<$0.5m \\
    fMoW~\cite{fmow} & 74$\times$58$\sim$16184$\times$16288 & 0.5m \\
    CVACT~\cite{cvact} & 1200 & 0.12m \\
    LoveDA~\cite{loveda} & 1024 & 0.3m \\
    \bottomrule
  \end{tabular}
  }
\caption{The six OOD image pools used in \textcolor{Gcolor}{G}\textcolor{Rcolor}{R}\textcolor{Acolor}{A}\textcolor{Scolor}{S}\textcolor{Pcolor}{P}-1k.}
\label{tab:dataset_summary}
\end{table}

To ensure data quality, we apply the BRISQUE~\cite{brisque} (Blind/Referenceless Image Spatial Quality Evaluator) metric to assess image clarity, contrast, and distortion. BRISQUE computes quality scores based on deviations from natural scene statistics. It is formally defined as:
\begin{equation}
    \text{BRISQUE}(I) = f\left(\phi(I)\right),
\end{equation}
where $\phi(I)$ denotes a set of natural scene statistics features extracted from locally normalized luminance patches of image $I$, and $f(\cdot)$ is a learned regression model that maps these features to perceptual quality scores. Lower scores indicate better perceived quality.

Following this, we discard all images with BRISQUE scores greater than or equal to 50:
\begin{equation}
    \mathcal{I}_\text{clean} = \{ I_i \mid \text{BRISQUE}(I_i) < 50 \}.
\end{equation}
As shown in Fig.~\ref{fig:grasp_bench}, images with high scores typically lack visual clarity or contain limited semantic information, making them unsuitable for constructing reasoning-intensive prompts.

For the remaining high-quality images, we employ Gemini-2.5-Pro~\cite{gemini} to generate reasoning-oriented prompts. Each data point consists of three elements: 1) a visually grounded question with logical reasoning requirements, 2) a directional answer referencing specific spatial locations, and 3) a detailed reasoning chain. Human annotators then identify positive points based on the answer and apply SAM2~\cite{sam2} to generate corresponding segmentation masks. 
For ambiguous or invalid cases, samples are discarded. In rare instances where SAM2 fails to produce satisfactory results, 
manual segmentation is performed using LabelMe~\cite{labelme}. All manually annotated cases undergo cross-validation by two independent annotators, and only those achieving full agreement are retained to ensure annotation correctness.

As a result, we construct \textcolor{Gcolor}{G}\textcolor{Rcolor}{R}\textcolor{Acolor}{A}\textcolor{Scolor}{S}\textcolor{Pcolor}{P}-1k, a high-quality benchmark of 1071 samples based entirely on OOD imagery, offering a rigorous evaluation for geospatial pixel reasoning.

\section{Experiments}\label{sec:Experiments}
\subsection{Experimental Settings}
\paragraph{Implementation Details.}
We adopt the Qwen2.5-VL-7B-Instruct~\cite{qwenvl25} as our base reasoning model and the SAM2-Large~\cite{sam2} as our base segmentation model. During both training and inference stages, we provided Qwen2.5-VL with the user-defined prompt illustrated in Fig.~\ref{fig:prompt} to guide its output. \textcolor{Gcolor}{G}\textcolor{Rcolor}{R}\textcolor{Acolor}{A}\textcolor{Scolor}{S}\textcolor{Pcolor}{P} is trained with the verl~\cite{verl} reinforcement learning framework, which provides a scalable implementation of GRPO. The training is performed on 8 NVIDIA A100-40G GPUs, with hyperparameters configured as follows: the KL loss coefficient is $5.0\times 10^{-3}$, the learning rate is $1.0\times 10^{-6}$, and the micro-batch sizes per device are $8$ for policy updates and $4$ for experience collection. Training runs for one epoch over the constructed dataset.

\paragraph{Evaluation Datasets.}
Our method is trained on the EarthReason~\cite{segearth} and GeoPixInstruct~\cite{geopix} training sets, and we merge their corresponding test sets to form a single in-domain evaluation dataset. In addition, we construct a completely out-of-domain benchmark, \textcolor{Gcolor}{G}\textcolor{Rcolor}{R}\textcolor{Acolor}{A}\textcolor{Scolor}{S}\textcolor{Pcolor}{P}-1k, from a newly curated image pool, which serves as our OOD test set.

\paragraph{Evaluation Methods.}
We compare our model with both natural image reasoning segmentation models (LISA-7b~\cite{lisa} and PixelLM-7b~\cite{pixellm}) and remote sensing referring image segmentation models (LGCE~\cite{lgce} and RMSIN~\cite{rmsin}), as well as three recent models featuring geospatial pixel reasoning (GeoPixel~\cite{geopixel}, GeoPix~\cite{geopix}, and SegEarth-R1~\cite{segearth}). To ensure a fair comparison, all baselines are fine-tuned or trained via SFT on the same training dataset as ours.

\paragraph{Evaluation Metrics.}
According to previous works, we evaluate our model using four widely adopted metrics: mIoU, gIoU~\cite{giou} and cIoU~\cite{ciou}. mIoU measures the mean Intersection-over-Union across all samples, reflecting overall segmentation accuracy. gIoU extends IoU by penalizing non-overlapping predictions via the inclusion of the smallest enclosing box area. cIoU further improves gIoU by incorporating the distance between the centers of predicted and ground-truth boxes as well as aspect ratio consistency. 

\begin{figure}[!t]
\centering
\includegraphics[width=0.8\columnwidth]{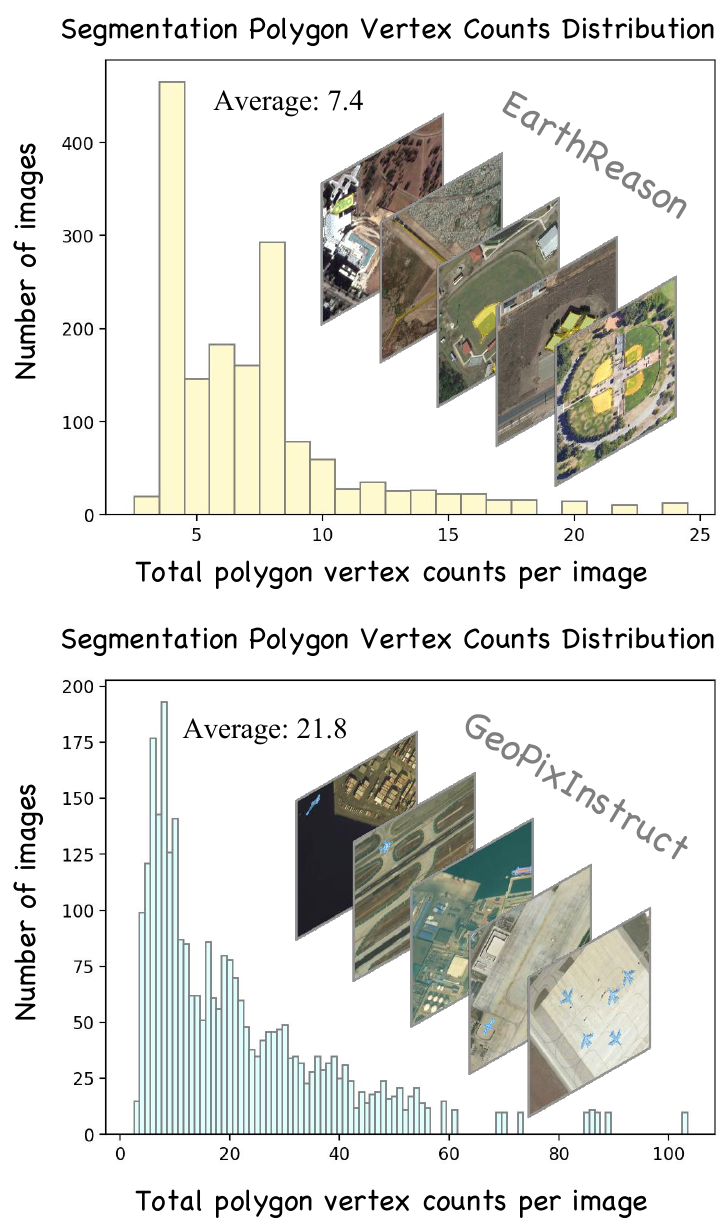}
\caption{
Distribution of polygon vertex counts in the training sets of EarthReason (top) and GeoPixInstruct (bottom). A higher number of vertices corresponds to more complex polygon boundaries, which directly increases annotation difficulty.
}
\label{fig:polygon_count}
\end{figure}

\subsection{Annotation Efficiency}  

To rigorously assess the efficiency gain from replacing dense masks with bounding boxes and points, we conduct two complementary analyses. The first measures practical annotation speed through a controlled user study. Ten computer science students familiar with LabelMe~\cite{labelme} are recruited as annotators. A total of 100 images are randomly selected, with 50 drawn from EarthReason and 50 from GeoPixInstruct. Five annotators are assigned to perform dense polygon segmentation, while the remaining five annotate bounding boxes and positive points. Each annotator works continuously without interruption, and the average annotation time is recorded. The results show that dense segmentation requires approximately 30 seconds per image, whereas bounding-box and points labeling requires only about 3 seconds, representing nearly a nine-fold increase in annotation efficiency  

Second, we analyze annotation complexity from the perspective of polygon vertex counts, where a larger number of vertices indicates greater annotation difficulty. We convert the dense masks from EarthReason and GeoPixInstruct into polygonal representations by setting the simplification parameter $\epsilon$ to $0.02$. This parameter controls the tolerance used in the polygonal approximation: larger values of $\epsilon$ produce polygons with fewer vertices by allowing greater deviation from the true contour. It is worth noting that $\epsilon=0.02$ is relatively large, meaning the resulting polygons are highly simplified compared to the original boundaries. We then compute the vertex count distribution for each dataset, as shown in Fig.~\ref{fig:polygon_count}. EarthReason exhibits relatively simple object boundaries with an average of 7.4 vertices per polygon, whereas GeoPixInstruct contains more complex shapes requiring an average of 21.8 vertices. In contrast, our BoP-based scheme represents each instance with only four points in total, since an axis-aligned bounding box can be uniquely defined by two diagonal vertices, which together with two positive points yield the complete representation. Even when compared with the simpler EarthReason annotations, our design still reduces the number of required points by approximately 46\%, demonstrating a substantial simplification of annotation complexity.

The results from both the user study and the polygon analysis demonstrate that our design significantly reduces annotation cost. At the same time, it still provides sufficiently informative signals for reward computation in geospatial pixel reasoning, as we further validate through quantitative and qualitative experiments in Sec.~\ref{subsec:qualitative}, Sec.~\ref{subsec:in_domain}, and Sec.~\ref{subsec:ood}.

\subsection{Qualitative Results}
\label{subsec:qualitative}

Our model is capable of generating both detailed reasoning chains and fine-grained segmentation masks. Benefiting from the strong reasoning capacity of Qwen2.5-VL and the pretrained segmentation ability of SAM2, we did not employ any SFT with segmentation or reasoning data. Instead, training is conducted entirely under \textit{PRIME} paradigm, where pure RL optimizes the MLLM with \textit{BoP-Rewards}. This paradigm activates the inherent potential of the cascaded architecture and leads to strong performance in both in-domain and OOD scenarios.

As shown in Fig.~\ref{fig:good_cases}, for the in-domain test set, the first row shows that our model can reliably distinguish the target object from multiple visually similar candidates, while also accurately identifying their relative positions and scale differences. The second and third rows further illustrate the model’s ability to handle reasoning-intensive queries. In these cases, it generates coherent and detailed reasoning chains and produces precise segmentation masks consistent with the natural-language instructions.  

For the OOD test set, the fourth row shows that the model can effectively localize small objects even under cluttered backgrounds, evidencing strong generalization to unseen domains. The fifth row presents an especially challenging reasoning problem, where the model outputs a well-structured explanation and a precise segmentation mask that aligns with the inferred conclusion. In addition, the sixth row highlights the model’s ability to tackle tasks requiring remote sensing knowledge. 

\subsection{Comparisons on In-domain Datasets}
\label{subsec:in_domain}

\renewcommand{\arraystretch}{1.5} 
\begin{table}[htbp]
    \centering
    \resizebox{\columnwidth}{!}{
    \begin{tabular}{lcccc}
        \toprule
        \textbf{Model} & \textbf{Pub.} & \textbf{mIoU} & \textbf{gIoU} & \textbf{cIoU} \\
        \midrule
        LISA-7b~\cite{lisa}          & CVPR'24  & 0.38  & 0.40  & 0.40 \\
        PixelLM-7b~\cite{pixellm}    & CVPR'24  & 0.37  & 0.37  & 0.37 \\
        LGCE~\cite{lgce}             & TGRS'24  & 0.29  & 0.29  & 0.29 \\
        RMSIN~\cite{rmsin}           & CVPR'24  & \textbf{0.50}  & 0.45  & 0.42 \\
        GeoPixel~\cite{geopixel}     & ICML'25  & 0.30  & 0.37  & 0.38 \\
        GeoPix~\cite{geopix}         & GRSM'25  & 0.41  & 0.42  & 0.39 \\
        SegEarth-R1~\cite{segearth}  & arxiv'25 & \uline{0.46}  & \uline{0.46}  & \uline{0.47} \\
        \textcolor{Gcolor}{G}\textcolor{Rcolor}{R}\textcolor{Acolor}{A}\textcolor{Scolor}{S}\textcolor{Pcolor}{P}(Ours)
        & --     & \uline{0.46}    & \textbf{0.48}        & \textbf{0.48} \\
        \bottomrule
    \end{tabular}
    }
    \caption{In-domain performance comparison.}
    \label{tab:in-domain compare}
\end{table}

As illustrated in Fig.~\ref{fig:qualitative comparison.}, the top four rows provide a qualitative comparison between our model and seven SOTA baselines on in-domain datasets. Most models are able to localize the relevant regions; however, our model produces the most fine-grained segmentation boundaries. This improvement arises from our precise bounding box and point predictions, which provide high-confidence prompts for SAM2 and thus lead to more accurate segmentation results. Tab.~\ref{tab:in-domain compare} presents the quantitative comparison, where our method achieves overall SOTA performance. In particular, it attains the best scores on both gIoU and cIoU, metrics that explicitly capture spatial deviation and boundary alignment between predictions and GT. These gains indicate that our framework yields more precise localization and geometrically consistent segmentations, underscoring the effectiveness of \textit{BoP-Rewards} in guiding fine-grained spatial reasoning. We note that RMSIN achieves the highest mIoU score. This outcome can be attributed to the composition of the in-domain benchmark, which includes a large proportion of referring image segmentation tasks. RMSIN is specifically designed for this setting, incorporating multi-scale feature interaction and orientation-robust convolution modules that allow it to better fit dense segmentation masks. Nevertheless, the superior gIoU and cIoU of our method highlight its stronger capability in reducing prediction deviations, thereby delivering more robust and reliable segmentation outcomes.

\subsection{Comparisons on \textcolor{Gcolor}{G}\textcolor{Rcolor}{R}\textcolor{Acolor}{A}\textcolor{Scolor}{S}\textcolor{Pcolor}{P}-1k}
\label{subsec:ood}
As shown in the bottom six rows of Fig.~\ref{fig:qualitative comparison.}, our model achieves substantially superior performance on the OOD test set compared with all other baselines. In the fifth row, our model successfully identifies trains from a bird’s-eye perspective and produces precise segmentation masks. The sixth and seventh rows present even more challenging cases: they involve not only reasoning-intensive queries but also GT objects hidden within extremely dense urban scenes. In these examples, only our model is able to accurately localize the targets, whereas other methods either output large coarse regions, fail to generate masks, or produce erroneous predictions due to low segmentation confidence. The final three rows highlight remote sensing–specific objects such as lakes, migration pathways, and cloud cover. Even under queries with high reasoning complexity, our model consistently localizes these objects and produces fine-grained segmentation masks. Notably, in the last two cases, our model produces more detailed and accurate boundaries than the provided GT. This suggests that \textit{PRIME} with \textit{BoP-Rewards} can elicit finer spatial reasoning from pretrained models, in some cases surpassing the quality of human annotations.  Tab.~\ref{tab:out-of-domain compare} further validates these findings with quantitative results: our model achieves improvements of 39\% in mIoU, 54\% in gIoU, and 39\% in cIoU on the OOD test set. These results highlight the robustness gains from both the \textit{PRIME} training paradigm and our carefully designed \textit{BoP-Rewards}.

\renewcommand{\arraystretch}{1.5} 
\begin{table}[htbp]
    \centering
    \resizebox{\columnwidth}{!}{
    \begin{tabular}{lcccc}
        \toprule
        \textbf{Model} & \textbf{Pub.} & \textbf{mIoU} & \textbf{gIoU} & \textbf{cIoU} \\
        \midrule
        LISA-7b~\cite{lisa}          & CVPR'24  & 0.25  & 0.16  & 0.14 \\
        PixelLM-7b~\cite{pixellm}    & CVPR'24  & 0.29  & 0.18  & 0.20 \\
        LGCE~\cite{lgce}             & TGRS'24  & 0.12  & 0.06  & 0.08 \\
        RMSIN~\cite{rmsin}           & CVPR'24  & 0.29  & 0.14  & 0.16 \\
        GeoPixel~\cite{geopixel}     & ICML'25  & 0.29  & 0.10  & 0.16 \\
        GeoPix~\cite{geopix}         & GRSM'25  & \uline{0.33}  & \uline{0.26}  & \uline{0.28} \\
        SegEarth-R1~\cite{segearth}  & arxiv'25 & 0.28  & 0.12  & 0.17 \\
        \textcolor{Gcolor}{G}\textcolor{Rcolor}{R}\textcolor{Acolor}{A}\textcolor{Scolor}{S}\textcolor{Pcolor}{P}(Ours)
        & --     & \textbf{0.46}    & \textbf{0.40}        & \textbf{0.39} \\
        \bottomrule
    \end{tabular}
    }
    \caption{Out-of-domain performance comparison.}
    \label{tab:out-of-domain compare}
\end{table}

\subsection{Ablation Study}


\begin{table}[htbp]
    \centering
    \large
    \renewcommand{\arraystretch}{1.2}
    \resizebox{\columnwidth}{!}{
    \begin{tabular}{l@{\hskip 15pt}ccc@{\hskip 18pt}ccc}
        \toprule
         & \multicolumn{3}{c}{\textbf{In-Domain}} & \multicolumn{3}{c}{\textbf{Out-of-Domain}} \\
        \midrule
         \textbf{Model} & mIoU & gIoU & cIoU & mIoU & gIoU & cIoU \\
        \midrule
        \textcolor{Gcolor}{G}\textcolor{Rcolor}{R}\textcolor{Acolor}{A}\textcolor{Scolor}{S}\textcolor{Pcolor}{P}-Zero  
        & 0.15  & 0.14  & 0.14 & 0.11  & 0.09  & 0.08 \\
        \textcolor{Gcolor}{G}\textcolor{Rcolor}{R}\textcolor{Acolor}{A}\textcolor{Scolor}{S}\textcolor{Pcolor}{P}-SFT   
        & \textbf{0.51}  & \textbf{0.50}  & \textbf{0.50} & 0.35  & 0.32  & 0.30 \\
        \textcolor{Gcolor}{G}\textcolor{Rcolor}{R}\textcolor{Acolor}{A}\textcolor{Scolor}{S}\textcolor{Pcolor}{P}-PRIME    
        & 0.46  & 0.48  & 0.48 & \textbf{0.46}  & \textbf{0.40}  & \textbf{0.39} \\
        \bottomrule
    \end{tabular}}
    \caption{
    Ablation study comparing \textit{SFT} and \textit{PRIME} training paradigms. \textit{SFT} achieves strong in-domain performance by leveraging fine-grained mask supervision, but this comes at the cost of substantially higher annotation requirements. In contrast, \textit{PRIME}, trained only with cost-efficient \textit{BoP-Rewards}, attains comparable in-domain scores while significantly surpassing \textit{SFT} on OOD benchmarks.
    }
    \label{tab:ablation_sft_rl}
\end{table}

\paragraph{SFT vs. PRIME.}
We compare SFT and \textit{PRIME} within our cascaded framework, as shown in Tab.~\ref{tab:ablation_sft_rl}. \textcolor{Gcolor}{G}\textcolor{Rcolor}{R}\textcolor{Acolor}{A}\textcolor{Scolor}{S}\textcolor{Pcolor}{P}-Zero refers to an untrained cascaded framework in which Qwen2.5-VL-7B-Instruct directly predicts bounding boxes from the image and instruction, which are then fed into SAM2 for mask generation. This baseline performs poorly for two main reasons: 1) the absence of output format regularization often leads to unparsable results; 2) Qwen2.5-VL itself is primarily optimized for natural images, leaving a significant domain gap when applied to remote sensing data.  

\textcolor{Gcolor}{G}\textcolor{Rcolor}{R}\textcolor{Acolor}{A}\textcolor{Scolor}{S}\textcolor{Pcolor}{P}-SFT follows the standard paradigm of prior works: Qwen2.5-VL produces a \texttt{[SEG]} token that is projected through a one-layer MLP into SAM2’s mask decoder, and both Qwen2.5-VL and SAM2 are jointly fine-tuned with dense mask supervision. This design yields strong in-domain performance for three main reasons. First, SFT has access to fine-grained segmentation masks as GT supervision, which are more informative than sparse box–point annotations, though this comes with a substantial annotation burden. Second, the representation space of a \texttt{[SEG]} token is inherently richer than that of box–point prompts, providing a higher theoretical performance ceiling, albeit at the cost of increased training complexity. Third, SFT is well suited to fitting the training data distribution, resulting in strong in-domain accuracy. However, this strength simultaneously limits its generalization ability.  

It should be emphasized that \textbf{this comparison is not entirely fair}: while SFT benefits from dense mask supervision, our \textit{PRIME} framework relies solely on bounding boxes and points. Nevertheless, under this stricter setting, \textcolor{Gcolor}{G}\textcolor{Rcolor}{R}\textcolor{Acolor}{A}\textcolor{Scolor}{S}\textcolor{Pcolor}{P}-PRIME \textbf{still achieves remarkable gains} in OOD evaluations ($+0.11$, $+0.08$, $+0.09$), demonstrating that \textit{PRIME} with \textit{BoP-Rewards} substantially enhances generalization. At the same time, its in-domain performance remains comparable to that of SFT, with only marginal reductions ($-0.05$, $-0.02$, $-0.02$ across mIoU, gIoU, and cIoU). These results underscore the practicality of our approach, as it achieves stronger robustness while relying on annotations that are far cheaper and easier to scale in real-world scenarios.

\begin{figure}[!t]
\centering
\includegraphics[width=1.0\columnwidth]{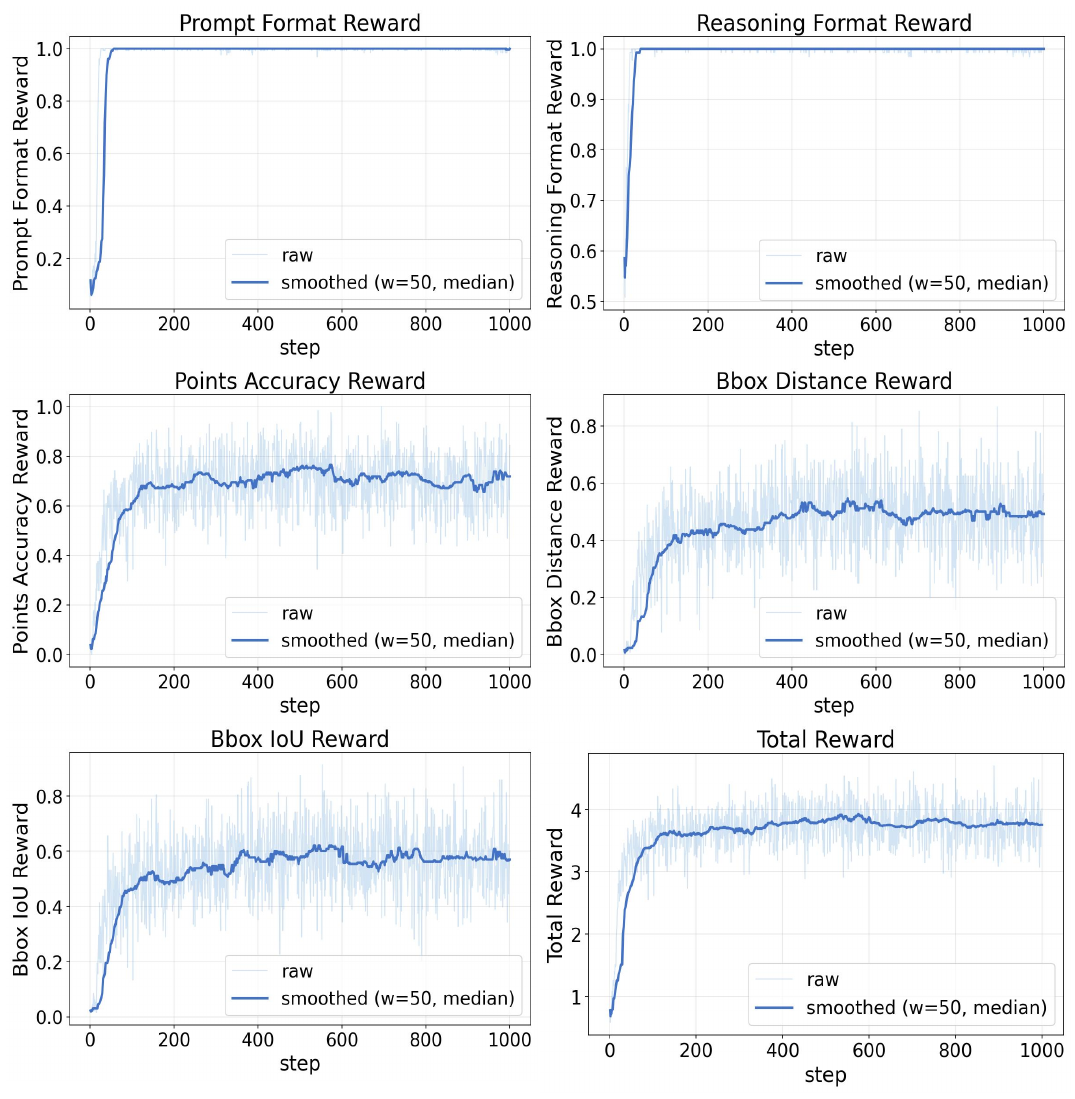}
\caption{
Reward curves for \textit{BoP-Rewards}. We report the first 1000 training steps across all five reward components as well as the total reward. The dark blue lines denote median-smoothed trends with a window size of 50, while the light blue traces show the raw rewards.
}
\label{fig:bop_rewards_curve}
\end{figure}

\paragraph{Ablation of BoP-Rewards.}

\newcommand{\cmark}{\checkmark}
\newcommand{\xmark}{\(\times\)}
\begin{table}[htbp]
  \large
  \centering
  \renewcommand{\arraystretch}{1.1} 
  \resizebox{\columnwidth}{!}{
  \begin{tabular}{lccccc}
    \toprule
    \textbf{Format} & \textbf{Bbox IoU} & \textbf{Bbox Dis} & \textbf{Points} & \textbf{mIoU$_{\text{id}}$} & \textbf{mIoU$_{\text{ood}}$} \\
    \midrule
     \xmark & \cmark & \cmark & \cmark & 0.20 & 0.11 \\
     \cmark & \xmark & \xmark & \cmark & 0.36 & 0.29 \\
     \cmark & \cmark & \xmark & \xmark & 0.42 & 0.40 \\
     \cmark & \cmark & \cmark & \xmark & 0.44 & 0.43 \\
     \cmark & \cmark & \cmark & \cmark & \textbf{0.46} & \textbf{0.46} \\
    \bottomrule
  \end{tabular}}
  \caption{Ablation of \textit{BoP-Rewards} design.}
  \label{tab:bbox_points_ablation}
\end{table}

We conduct a detailed ablation study on our \textit{BoP-Rewards} design, as reported in Tab.~\ref{tab:bbox_points_ablation}. The format reward includes both reasoning and prompt format reward, the Bbox IoU reward provides strict overlap supervision, the Bbox distance reward penalizes deviations in box localization, and the point accuracy reward evaluates the correctness of predicted points. As shown in the first row of Table~\ref{tab:bbox_points_ablation}, since properly structured outputs are a prerequisite for computing the other rewards, the model fails to learn effectively without the format reward. A comparison between the second and third rows shows that box-based accuracy rewards are more effective than point-based ones. This can be attributed to the pretrained model’s stronger prior for generating bounding boxes, as well as to the fact that boxes provide a more informative prompt for SAM2. The last four rows further demonstrate that each reward contributes positively: all designs improve both in-domain and OOD performance, and the combination of all of them achieves the best results with 0.46 mIoU in both settings.  

To better understand how these rewards behave during training, we further plot their trajectories in Fig.~\ref{fig:bop_rewards_curve} under the setting where all five rewards are applied. The two format rewards converge rapidly within the first few steps, indicating that structural correctness is quickly established and serves as an early guiding signal for subsequent learning. In contrast, the accuracy-related rewards evolve more gradually but steadily, reflecting the progressive refinement of localization precision. Importantly, the total reward exhibits a clear upward and stabilizing trend, confirming that the overall training process is stable and that the reward components work in a complementary manner.

\section{Conclusion}\label{sec:Conclusion}
This work tackles the task of geospatial pixel reasoning, which requires generating fine-grained segmentation masks in remote sensing imagery from natural-language instructions. Prior approaches have relied on dense mask supervision and SFT, which incur prohibitive annotation costs and exhibit poor generalization beyond the training distribution.  

We proposed \textcolor{Gcolor}{G}\textcolor{Rcolor}{R}\textcolor{Acolor}{A}\textcolor{Scolor}{S}\textcolor{Pcolor}{P}, a structured policy-learning framework that cascades an MLLM with a pretrained segmentation model. Central to this framework are two components: \textit{PRIME}, an RL paradigm that replaces SFT to more effectively align reasoning and grounding with task objectives, and \textit{BoP-Rewards}, a cost-aware reward design that substitutes dense masks with bounding boxes and points while preserving informative training signals.  

To enable systematic evaluation, we trained \textcolor{Gcolor}{G}\textcolor{Rcolor}{R}\textcolor{Acolor}{A}\textcolor{Scolor}{S}\textcolor{Pcolor}{P} and all strong baselines on EarthReason and GeoPixInstruct, and constructed a unified in-domain benchmark from their test sets. In addition, we released \textcolor{Gcolor}{G}\textcolor{Rcolor}{R}\textcolor{Acolor}{A}\textcolor{Scolor}{S}\textcolor{Pcolor}{P}-1k, a new OOD benchmark featuring reasoning-intensive queries, reasoning traces, and fine-grained masks. Experiments demonstrate that our method not only attains SOTA performance in-domain but also achieves up to a 54\% improvement under distribution shift.  

In summary, this study shows that RL with structured, cost-effective rewards offers a practical and robust paradigm for bridging vision–language reasoning and dense prediction in remote sensing, thereby paving the way for more scalable and generalizable geospatial analysis.  





\section*{Acknowledgments}

This work is supported by the National Natural Science Foundation of China under Grant 62206321.

{
	\begin{spacing}{1.17}
		\normalsize
		\bibliography{ISPRSguidelines_authors} 
	\end{spacing}
}

\end{document}